\ifdefined\pdfobjcompresslevel
\fi
\PassOptionsToPackage{table,dvipsnames}{xcolor}
\documentclass[]{selfevolagent}

\usepackage[utf8]{inputenc}
\usepackage[T1]{fontenc}
\usepackage{amsfonts}
\usepackage{amsmath}
\usepackage{amssymb}
\usepackage{mathtools}
\usepackage{float}

\newcommand{\ourmethod}{AMID}


\title{Towards Autonomous and Auditable Medical Imaging Model Development}
\author[1,*]{Shengyuan Liu}
\author[1,*]{Jia-Xuan Jiang}
\author[1]{Boyun Zheng}
\author[1]{Cheng Wang}
\author[2]{Zipei Wang}
\author[1]{Wentao Pan}
\author[1]{Hongtao Wu}
\author[5]{Houwen Peng}
\author[3]{Yu Gu}
\author[4,\dagger]{Lichao Sun}
\author[1,\dagger]{Yixuan Yuan}

\affiliation[1]{The Chinese University of Hong Kong}
\affiliation[2]{Institute of Automation, Chinese Academy of Sciences}
\affiliation[3]{Microsoft Research}
\affiliation[4]{Lehigh University}
\affiliation[5]{Independent Researcher}

\contribution[*]{Equal contribution}
\contribution[\dagger]{Corresponding author}

\abstract{%

Large language model (LLM) agents are beginning to automate machine learning engineering (MLE) by coupling planning, code execution, debugging, and empirical feedback. Translating this capability to medical imaging remains difficult because each task imposes modality-specific experimentation and strict requirements for validation protocols and prediction artifacts. Here we introduce \ourmethod{}, an autonomous multi-agent framework for medical imaging model development. \ourmethod{} first proposes Data-Conditioned Method Planning, which refines coarse task-level search spaces into executable, parallelizable method lanes grounded in task-specific data analysis and runnable medical-imaging resources. It then develops Verification-Guided Two-Stage Optimization, moving from broad early exploration of diverse method lanes to selective exploitation of promising candidates while enforcing strict verification of validation protocols, metric computation, and prediction artifacts throughout the optimization. Across 20 medical imaging challenge tasks spanning diverse modalities and prediction types, \ourmethod{} outperformed evaluated general-purpose MLE systems and, on several tasks, approached or matched strong human-designed challenge solutions. These results suggest that \ourmethod{} can turn task-specific medical imaging model development from bespoke manual engineering into an agentic workflow for producing high-performing and auditable model artifacts across heterogeneous tasks.

\par\vspace{-0.35\baselineskip}
{\small\itshape Note: This is an ongoing preliminary technical report.}
}

\projectpage{https://github.com/CUHK-AIM-Group/AMID}

\begin{document}

\maketitle

\begin{figure}[!h]
\centering
\includegraphics[width=\textwidth]{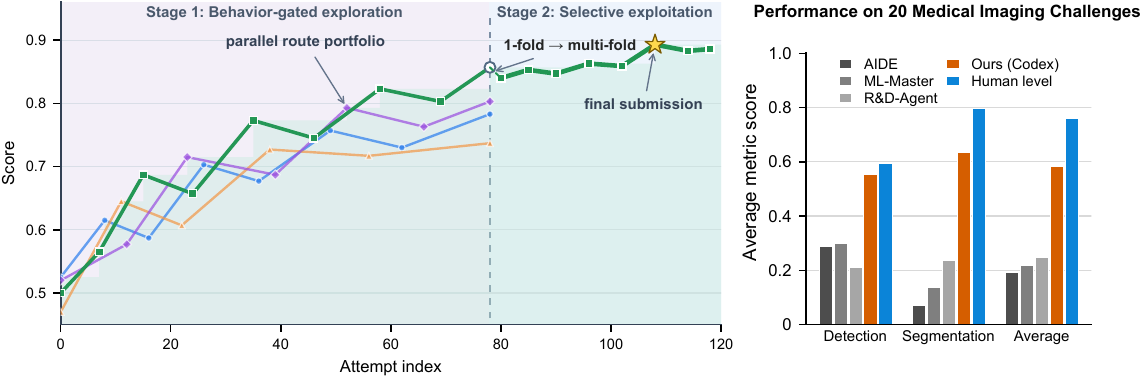}
\caption{Overview of \ourmethod{}'s verification-guided optimization and benchmark performance. The two-stage optimization moves from broad exploration of parallel method lanes to selective exploitation of promoted candidates, while reviewer checks verify validation protocols, metric computation, and prediction artifacts throughout the optimization; shaded regions mark the two stages, and the star marks the final selected submission. The category-level bar chart reports average metric scores over 20 ReX-MLE medical imaging challenges.}
\label{fig:main_overview}
\end{figure}
\FloatBarrier

\section{Introduction}
\label{sec:introduction}
Large language model (LLM) agents are emerging as a promising paradigm for automating machine learning engineering (MLE), shifting model development from one-shot code generation toward iterative, feedback-driven experimentation~\citep{aygun2026ai,chan2024mle,jimenez2024swebench}. In this paradigm, agents iteratively transform methodological ideas into executable experiments, using failures, validation signals, and accumulated evidence to guide subsequent refinements. Existing MLE agents~\citep{Jiang2025AIDEAE,liu2025ml,yang2025r,du2025automlgen,du2026mlevolve} have made this loop more structured and feedback-aware, coordinating exploration, execution, and evaluation while separating high-level methodological reasoning from code-level implementation and repair. Autonomous research agents extend this perspective beyond model development, showing that empirical loops must ultimately support scientific claims rather than merely produce improved results~\citep{lu2024aiscientist,yamada2025aiscientistv2,huang2025deepscientist,yang2026aris,li2026autosota,li2024autoresearchclaw}. Together, these developments suggest that agentic MLE can substantially reduce the engineering burden of model development, but only when its search process is executable and auditable.

Translating agentic MLE to medical imaging is not simply a matter of applying a generic experimentation loop to clinical datasets \citep{huang2026autonomous,ferber2026towards}. Medical imaging model development couples machine learning decisions with clinical and data-specific constraints, where the choice of preprocessing, model family, validation protocol, metric, and prediction artifact is shaped by the imaging modality, clinical objective, annotation process, and patient-level data organization~\citep{litjens2017survey,ronneberger2015unet,isensee2021nnu}. As a result, an apparently standard prediction task often requires domain-conditioned engineering rather than generic pipeline search. Biomedical foundation models and generalist medical AI systems, including Med-PaLM M~\citep{tu2023generalist}, BiomedGPT~\citep{zhang2024biomedgpt}, BioMedParse~\citep{zhao2025foundation}, MedVersa~\citep{zhou2026medversa}, and MedSAM~\citep{ma2024medsam}, provide reusable representations, segmentation priors, and multimodal reasoning capabilities, but they remain building blocks within a broader development workflow. They do not by themselves determine how a particular dataset should be preprocessed, modeled, evaluated, and adapted to its clinical objective~\citep{degrave2021shortcut,geirhos2020shortcut,gichoya2022ai,yang2024limits,ong2024shortcut}. Thus, medical imaging requires autonomous MLE systems that can reason from task-specific data and domain constraints, rather than merely reuse general-purpose modeling routines.

This requirement exposes two limitations of generic autonomous MLE systems. First, the search space constructed by generic agents is often too coarse-grained for heterogeneous medical-imaging tasks. A task may be recognized as classification, segmentation, detection, or restoration, but the appropriate method family can depend on modality, anatomy, acquisition protocol, annotation semantics, clinical target, and submission constraints. Without explicitly converting these medical-imaging attributes into executable and parallelizable method lanes, an agent may spend compute on pipelines that are technically executable but medically inappropriate, poorly matched to the data, or unable to produce valid final artifacts. Second, experimental feedback in medical imaging is often expensive, delayed, and potentially unreliable. Unlike many CPU-oriented or lightweight benchmark tasks where feedback can be obtained quickly, medical-imaging experiments may require high GPU budgets, long training runs, memory-aware inference, and strict patient-level or site-level splits. Under such conditions, agents may be biased toward low-cost shortcut experiments, incomplete training, weak validation protocols, or invalid artifacts. A single validation score can therefore be unstable or misleading under small cohorts, patient-level dependence, scanner or site effects, leakage, metric mismatch, fold mistakes, or invalid prediction artifacts, motivating a strict verification scheme that prevents untrusted evidence from driving the search process.

To address these challenges, we introduce \textbf{\ourmethod{}}, an
\textbf{\underline{A}}utonomous multi-agent framework for
\textbf{\underline{M}}edical \textbf{\underline{I}}maging model
\textbf{\underline{D}}evelopment. Given a medical-imaging task definition and its associated data, \ourmethod{} aims to turn task-specific model development from labor-intensive manual engineering into an agentic workflow that produces high-performing, validated, and auditable model artifacts, including executable code, validation evidence, and submission-ready prediction artifacts. The framework is designed around a central premise: reliable medical MLE requires both explicit construction of the candidate method space and rigorous verification of the evidence used to select among candidates. First, \ourmethod{} performs Data-Conditioned Method Planning (DCMP), which refines coarse task-level search spaces into executable, parallelizable method lanes grounded in task-specific data analysis and runnable medical-imaging resources. Rather than searching over generic pipelines, DCMP analyzes the modality, label structure, evaluation target, and submission requirements of each task, then grounds candidate methods in relevant code, literature, prior templates, and foundation models, organizing them into method lanes with clear assumptions, implementation plans, and validation requirements. Second, \ourmethod{} performs Verification-Guided Two-Stage Optimization, a hierarchical optimization procedure that moves from broad early exploration of diverse method lanes to selective exploitation of promising candidates. Throughout both stages, a reviewer process enforces strict verification of validation protocols, metric computation, and prediction or submission artifacts before any attempt can drive promotion or final selection. We conduct a comprehensive evaluation of \ourmethod{} on 20 medical-imaging challenge tasks spanning diverse modalities and prediction types, demonstrating that \ourmethod{} outperforms evaluated general-purpose MLE systems and, on several tasks, approaches or matches strong human-designed challenge solutions. 

Our contributions can be summarized as:
\begin{itemize}
    \item To our knowledge, \ourmethod{} is the first autonomous multi-agent MLE framework designed for general medical imaging model development, producing auditable model packages with executable code, validation evidence, and submission-ready prediction artifacts across heterogeneous tasks and modalities.

    \item We propose Data-Conditioned Method Planning (DCMP), which refines coarse task-level search spaces into executable and parallelizable method lanes grounded in task-specific data analysis, relevant medical-imaging code, literature, prior templates, and foundation models.

    \item We introduce Verification-Guided Two-Stage Optimization, which moves from broad early exploration of diverse method lanes to selective exploitation of promising candidates while enforcing reviewer checks on validation protocols, metric computation, and prediction artifacts throughout the optimization process.
\end{itemize}

\section{System Overview}
\label{sec:system_overview}
\begin{figure}[t]
\centering
\includegraphics[width=\textwidth]{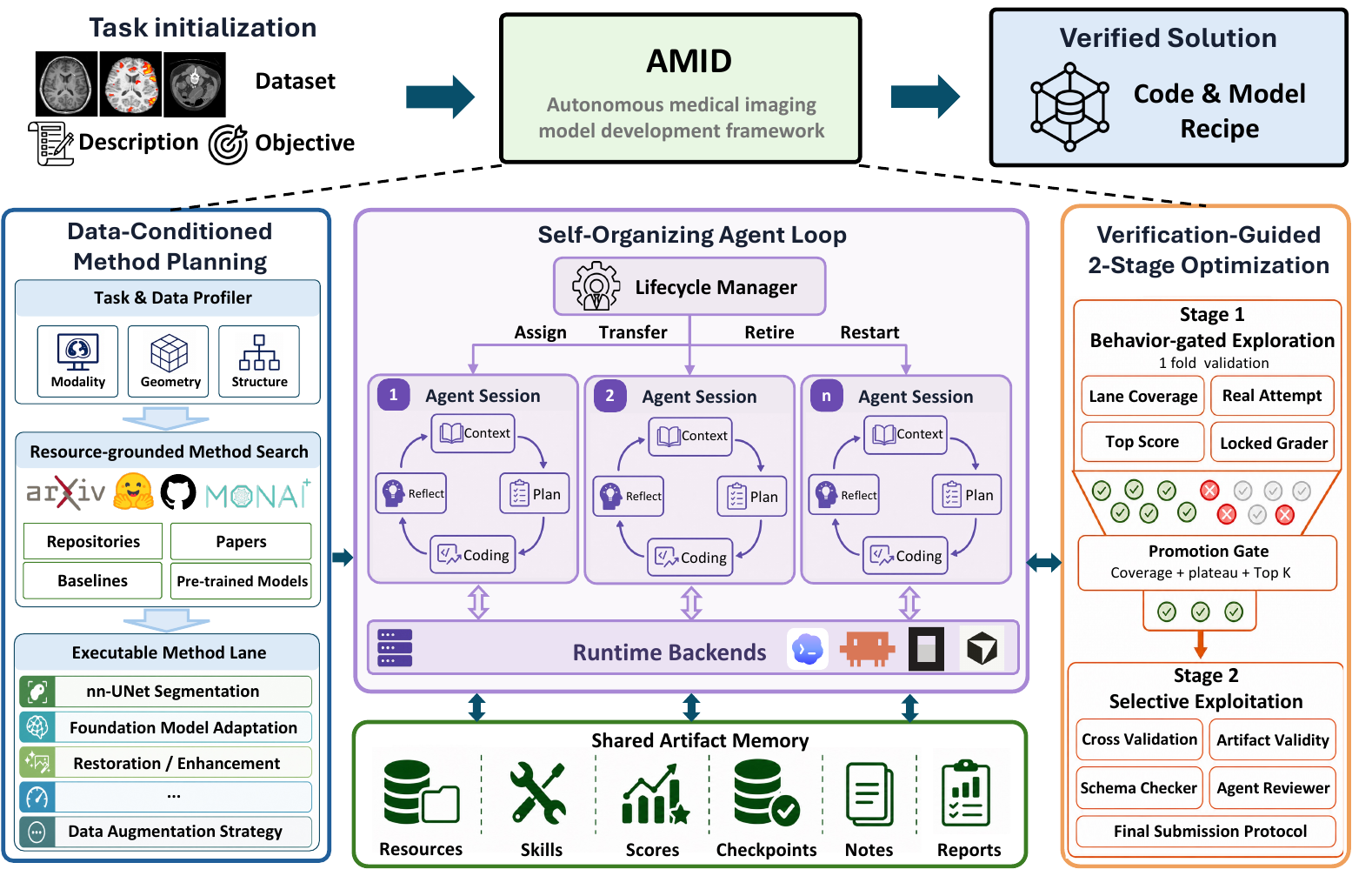}
\caption{Overview of \ourmethod{}. Starting from task initialization, \ourmethod{} profiles the medical imaging data and converts task evidence into executable method lanes through Data-Conditioned Method Planning. A self-organizing agent loop then explores a diverse method portfolio, performs selective exploitation on promising candidates, and accepts a final model package only when reviewer-checked artifacts satisfy the task contract.}
\label{fig:method_overview}
\end{figure}

As shown in Figure~\ref{fig:method_overview}, \ourmethod{} turns a medical-imaging task into an auditable model package through a data-conditioned and verification-controlled agent workflow. The system first establishes the task contract and profiles the input data, then converts the resulting evidence into executable method lanes, coordinates agents through a two-stage explore-and-exploit optimization, and accepts a final model or submission only when the supporting artifacts satisfy the task contract. This section follows the same system view: we first formalize the model-development problem, then describe how \ourmethod{} plans method lanes, optimizes them through Verification-Guided Two-Stage Optimization, and preserves evidence across agent runtimes.

\subsection{Task and Artifact Contract}

\ourmethod{} addresses a practical medical imaging model-development problem. The input is intentionally minimal: a medical-imaging dataset and a task definition that specifies the target output, evaluation metric, and, when applicable, the submission protocol. The dataset may contain 2D images, 3D volumes, pathology tiles, paired enhancement data, detection annotations, segmentation masks, class labels, graph labels, or image-quality scores. The task definition may come from a clinical modeling request or from a challenge description. The expected output is not a single text answer or a suggested architecture, but a complete model package: executable training and inference code, model weights or checkpoints, prediction files, validation scores, final submission artifacts when required, and an audit trail showing that the result was produced under the correct data, metric, split, and submission contract.

This framing treats autonomous medical imaging AI development as accountable engineering. A high local score is useful only if the score is comparable, the metric direction is correct, the split is legal, the prediction format matches the evaluator, and the model lineage can be inspected. \ourmethod{} therefore searches for model-development procedures that are both high-performing and verifiable. The system is designed to start from basic task information and end with an auditable candidate model, while making intermediate choices inspectable to human reviewers.

\subsection{Workflow Architecture}
\ourmethod{} organizes autonomous medical-imaging model building as a staged multi-agent run. The system first materializes a workspace from the task files and records the modeling target, evaluation metric, prediction schema, and required output contract. It then profiles the public data view and writes an evidence-backed summary of modality, dimensionality, supervision type, sample organization, geometry, label structure, and risks such as leakage, imbalance, sparse masks, or invalid output formats. Data-Conditioned Method Planning uses this profile to build a method-lane portfolio, grouping implementation-backed routes into method families for parallel exploration. Verification-Guided Two-Stage Optimization then turns this portfolio into an executable optimization process that moves from broad early exploration of diverse method lanes to selective exploitation of promising candidates. In both stages, reviewer checks verify the validation protocol, metric computation, and prediction or submission artifacts before an attempt can influence promotion, repair, or final selection. During exploration, workers cover different method lanes and produce comparable attempts with code, validation scores, logs, predictions, and submission artifacts. During selective exploitation, workers repair implementation issues, complete validation evidence, tune or ensemble selected variants when appropriate, and prepare the final deliverable from reviewer-accepted candidates.

This model-building workflow supports several deployment modes. In the default open-ended mode, agents continue improving candidate models until the user stops the run or accepts a result. \ourmethod{} also supports a fixed-time-budget mode for settings such as competitions, where the manager schedules exploration, promotion, and finalization under an explicit deadline. The final deliverable is configurable: a task may require a submission file, prediction files referenced by that submission, an executable model package such as a Dockerized predictor, or both. The same search policy can be executed by different coding-agent backends, including Codex, Claude Code, OpenCode, Cursor Agent, and Kiro, and mixed-backend runs can allocate different agents to different runtimes while preserving the same shared artifacts, validation protocol, and final-artifact checks.

\subsection{Data-Conditioned Method Planning}

Data-Conditioned Method Planning (DCMP) addresses a failure mode we observe in general-purpose autonomous model-building agents. Medical imaging tasks are rarely solved well by selecting a generic training recipe from the task statement alone. The same nominal problem type can require very different choices depending on modality, dimensionality, voxel spacing, annotation density, object scale, patient grouping, metric definition, and output format. Conversely, a broad literature or web search often returns impressive methods that are not executable in the current setting because the code is unavailable, the checkpoint is gated, the data interface is incompatible, or the method assumes a different imaging geometry. DCMP is designed to make planning conditional on the actual data and on resources that can be used by workers. It converts the task contract, public data evidence, and executable resource inventory into a portfolio of method lanes for search.

\paragraph{Task and data profiling.}

The first stage builds an evidence-backed profile of the public task view. \ourmethod{} inspects task descriptions, README files, sample submissions, metric definitions or grader-facing hints, visible training data, labels, metadata, and the required prediction artifacts. This phase does not train models or search for methods. Its purpose is to make the dataset legible for later planning and to prevent each worker from rediscovering the same constraints.

The profile records several groups of planning-relevant evidence. At the inventory level, it summarizes file types, directory layout, train/test organization, sample counts, identifier structure, pairing relationships, missing or duplicated cases, and visible strata such as organ, site, scanner, view, phase, stain, or acquisition protocol when these are available. At the image-representation level, it samples raw files to estimate shape, channel count, dtype, intensity range, spacing, orientation, affine consistency, anisotropy, and volume-size distribution. At the supervision level, it characterizes the target type and label statistics: class balance for classification, foreground ratio and connected components for segmentation, box counts and object-size ranges for detection, alignment and contrast shifts for paired restoration, and modality missingness or registration relationships for multimodal tasks. At the evaluation level, it records the primary metric, metric direction, fold or group constraints, patient-level leakage risks, submission columns, referenced prediction files, and format checks that must hold before a score is trusted.

These diagnostics become search predicates. For example, sparse 3D vessel masks suggest patch sampling, spacing normalization, topology-aware post-processing, and memory-aware inference; histopathology data suggest tiling, stain variation handling, and slide- or patient-level aggregation; paired ultrasound enhancement requires alignment audits before pixel losses are trusted. The output is therefore not only a descriptive report, but a set of constraints and opportunities that downstream method search can use.

\paragraph{Resource-grounded method search.}

The second stage searches for methods under those data conditions. \ourmethod{} distinguishes between two resource sources. The first is a pre-seeded local resource library that can be provided before a run. In our implementation, this library includes common medical-imaging toolkits such as nnU-Net~\citep{isensee2021nnu} and MONAI~\citep{cardoso2022monai}. nnU-Net provides a strong starting point for segmentation tasks that require task-specific dataset conversion, fold management, preprocessing, training, and inference. MONAI provides reusable medical-imaging components for data loading, spatial and intensity transforms, losses, network architectures, pretrained bundles, and deployment patterns across segmentation, classification, detection, restoration, and other tasks. \ourmethod{} treats these resources as optional implementation anchors rather than mandatory baselines: a resource is assigned to a lane only when the profiled modality, task family, data geometry, and output contract make it actionable.

The second source is a curated foundation-model and external-resource search. \ourmethod{} can consult a registry of medical-imaging foundation models indexed by domain, modality, task family, keywords, access status, license, model card, repository, and downloadable files. The registry includes, for example, promptable segmentation models such as MedSAM~\citep{ma2024medsam} and MedSAM2~\citep{ma2025medsam2}, anatomy-oriented tools such as TotalSegmentator~\citep{wasserthal2023totalsegmentator} and VISTA3D~\citep{he2025vista3d}, and domain-specific foundation models for settings such as pathology, chest radiography, CT, MRI, and retinal imaging. For each candidate, the planner checks whether it matches the profiled data and whether its usable artifacts can be verified or acquired, so workers receive resources that are relevant to the task rather than a generic list of popular models.

DCMP combines the data profile and resource evidence into a hybrid method graph. Nodes represent data conditions, method families, and implementation resources; edges explain why a method family is plausible under the observed task constraints. The final lane portfolio is broader than a single best guess. It may include a conservative nnU-Net or MONAI baseline, a foundation-model adaptation lane, a task-specific architecture lane, a preprocessing- or post-processing-heavy lane, an ensemble lane, and calibration-only lanes that verify data loading, metric computation, and submission mechanics. Primary and secondary lanes must be implementation-backed. Lower-confidence, mismatched, paper-only, or over-budget ideas are retained as fallback lanes, downranked, or excluded rather than silently removed.

Each method lane contains a modeling hypothesis, the data conditions that motivate it, required resources, implementation requirements, preprocessing assumptions, validation obligations, approximate training budget, expected artifacts, and likely failure modes. Worker agents therefore start search from executable routes rather than from open-ended brainstorming. This is the role of DCMP: it does not decide the final model, but it shapes the search space so that subsequent exploration spends effort on methods that are both medically plausible for the observed data and operationally runnable in the current workspace.

\subsection{Verification-Guided Two-Stage Optimization}

Medical-imaging model building often admits multiple qualitatively different solution families, and different families can dominate under different data conditions. A task may be approached through a robust nnU-Net-style pipeline, a foundation-model adaptation, a task-specific architecture, or a preprocessing/post-processing-heavy route. This diversity makes early commitment to a single direction risky. It also requires strict verification beyond score-driven optimization alone: a high early score may come from a narrow split, leakage, a metric or formatting mismatch, or an implementation that cannot produce the required final artifact. \ourmethod{} therefore uses Verification-Guided Two-Stage Optimization over the method-lane portfolio produced by DCMP. The optimization units are lanes grouped into method families, where each lane specifies a concrete modeling hypothesis, resource set, and validation obligation. The process moves from broad early exploration of diverse method lanes to selective exploitation of promising candidates. In both stages, a reviewer checks whether each submitted attempt satisfies the active validation protocol, computes the metric correctly, and produces valid prediction or submission artifacts before the evidence can drive promotion or final selection.

\paragraph{Stage 1: Behavior-gated Exploration.}
The first stage gives \ourmethod{} a portfolio-level view of the DCMP search space. Worker agents are distributed across method families and submit attempts tied to specific method lanes. Each attempt records the committed code, predictions or submission artifacts, validation score, metadata, and lane attribution. \ourmethod{} uses these records to estimate which lanes are feasible, which method families show empirical promise, and which failure modes recur across implementations. The goal is not to exhaust every possible variant, but to obtain enough comparable behavior from the major lanes before committing the search to a small set of candidates.

This exploration is behavior-gated by a reviewer that is independent of the worker agents. Workers can propose an implementation plan for a lane and submit code, but they cannot certify their own evidence. Each submission is first recorded as a pending attempt with its committed code, worker identity, lane metadata, shared-state checkpoint, declared budget class, and produced artifacts. The reviewer then evaluates the frozen commit outside the worker session, checking whether the implementation plan was actually carried out under the active protocol and whether the resulting evidence is admissible. These checks cover the validation split, metric direction and computation, complete predictions or submission files, non-duplicate payloads, required logs or checkpoints, and traceability from code to artifacts. Incomplete runs, malformed predictions, duplicate payloads, tune-only checks, and metric-incompatible submissions remain useful for diagnosis, but they are not counted as valid lane coverage. Only attempts that pass this independent review become real, attributable, and comparable evidence. Once required lanes have enough reviewer-accepted attempts and show plateau behavior or a documented waiver, the manager ranks lanes by their best valid evidence and promotes the strongest candidates, optionally preserving a close second lane as a challenger.

\paragraph{Stage 2: Selective Exploitation.}

The second stage concentrates computation on the promoted candidates. Workers no longer perform broad method discovery; they start from accepted or promoted anchors and convert them into complete, auditable model packages. Optimizer workers improve the leading recipe through model, training, inference, test-time augmentation, post-processing, or ensemble changes. Challenger workers preserve limited capacity for a distinct promoted lane when the first-stage evidence does not clearly identify a single winner. Repair workers handle candidates that have useful raw scores but fail protocol or artifact checks, while finalizer workers prepare the final prediction, submission, or model package from an accepted anchor.

The same reviewer gate remains active during selective exploitation. A stage-2 candidate must satisfy the active validation protocol and produce evidence that can be inspected after the run. For medical-imaging tasks, this includes correct metric direction, legal data access, locked-fold usage when applicable, complete fold outputs, correct mean-score computation, fold provenance, fold-invariant pipelines, valid prediction schema, complete required files, and traceability from the final package back to an accepted attempt. Candidates that fail these checks may still inform repair or further optimization, but they are not admitted to the accepted candidate set or selected as the final model. The final package is therefore chosen from attempts that are both high-performing and reviewer-verified.

\subsection{Self-Organizing Agent Loop}
\label{sec:agent_loop}

\ourmethod{} adopts a self-organizing execution substrate and specializes it for medical-imaging model construction. The goal is to let multiple coding agents search asynchronously without hiding evidence inside private chat histories. The substrate is backend-agnostic: agents implemented with Codex, Claude Code, OpenCode, Cursor Agent, Kiro, or mixed backends can participate in the same run. Each worker runs in an isolated git worktree, but all workers operate against the same task contract, method lanes, attempts, notes, skills, resources, and audit artifacts. Unlike a fully peer-to-peer agent society, however, \ourmethod{} places a central lifecycle manager around this substrate. Agents remain autonomous in implementation and debugging, but their creation, context, interruption, reassignment, retirement, and finalization are controlled by the manager.

\paragraph{Lifecycle manager.}

The lifecycle manager serves as the operational control plane of the agent loop. It maintains an explicit state for each worker, covering initialization, code editing, evaluation waiting, feedback reflection, reassignment, retirement, and artifact preparation. At launch time, the manager creates the worker workspace, attaches the shared state view, injects the task contract and current lane context, and starts the selected coding-agent backend. During execution, it monitors process health, recent attempts, idle time, context pressure, and signs of diminishing progress. When intervention is needed, the manager can interrupt and resume an existing session with a targeted prompt, rather than replacing it with a fresh conversation.

This lifecycle policy gives the system a recoverable outer structure. Worker sessions may fail, stall, exhaust their current direction, or approach context limits; in each case, the manager applies an explicit transition rather than leaving the population unmanaged. It can restart a session, backfill capacity, request documentation of a useful procedure, or shift a candidate-producing worker toward artifact completion. Thus, parallel agents are not merely launched and left to run independently. They are maintained as managed computational processes whose states, transitions, and responsibilities remain visible to the system.

\paragraph{Manager-mediated self-organization.}

Building on this lifecycle machinery, \ourmethod{} realizes self-organization as an evidence-driven allocation of research effort rather than as coordination among agents with fixed identities. The unit of allocation is a temporary exploration focus assigned to a worker session. Such a focus denotes a direction of investigation rather than a durable role: it may correspond to a conservative baseline, a foundation-model adaptation, a data-processing route, a post-processing route, or a finalization path. Because foci are temporary and state-dependent, the worker population can be continuously reconfigured as the run unfolds. The resulting organization is therefore not a static ensemble of specialists, but a mutable distribution of effort over competing directions of progress.

The worker loop in Figure~\ref{fig:method_overview} provides the local evidence used for this reconfiguration. Each worker turns shared context into a bounded implementation plan, tests that plan in its own worktree, and externalizes the outcome through a public reflection. The manager aggregates these public traces to decide how collective attention should be redistributed: productive directions can be reinforced, redundant efforts merged, saturated branches deprioritized, and exploratory threads preserved when diversity remains useful. This yields a two-level organization of search: workers exercise local autonomy over implementation, debugging, and evaluation, while the manager performs global, evidence-driven allocation of research attention.

\paragraph{Shared memory.}

The shared memory is a file-system state rather than an in-context transcript. \ourmethod{} stores attempt records, experiment notes, method-lane files, resource manifests, validation state, reviewer reports, final-artifact records, manager state, and reusable skills in a shared workspace visible to all agents. Attempt records bind a score to a commit hash, agent identity, validation protocol, method-lane attribution, artifact paths, and grader feedback. Notes record observations such as preprocessing failures, fold-specific behavior, resource blockers, or promising next experiments. Because this state is durable and shared, later workers can inspect what has already been tried, reuse successful code paths, avoid known failures, and audit why a candidate was promoted or rejected.

Skills are treated as a structured part of this memory. \ourmethod{} seeds each run with skills for organizing files, using reference repositories, reading papers for reproduction, creating new skills, and adapting foundation models. The foundation-model skill, for example, does not select a model by itself; it is invoked after DCMP or a worker chooses a specific model, checkpoint, or model-zoo bundle. It then verifies the local resource, reads the model card or configuration, maps the model interface to the current task, preserves the task-native data, fold, and evaluation pipeline, records provenance, and runs smoke tests before expensive training. Agents can also create or update skills when a repeated workflow becomes useful. In this way, shared memory stores not only results, but also procedures that allow successful behavior to diffuse across workers during the same run.

\paragraph{Heartbeat interventions.}

\ourmethod{} uses heartbeat prompts to keep long-running agents aligned with the search objectives. The manager monitors the attempt stream, per-agent progress, global evaluation count, stalled workers, and plateau behavior. When a heartbeat triggers, the manager interrupts and resumes the agent with a targeted prompt rather than resetting the session. Reflection heartbeats ask agents to convert recent results into experiment notes with concrete causes and next actions. Consolidation heartbeats ask agents to synthesize shared notes, identify contradictions, summarize team state, and expose uncovered or over-exploited directions. Plateau heartbeats force a pivot decision when an agent has stopped improving, requiring a public focus note that names the current focus, evidence, remaining budget, and abandon condition. In \ourmethod{}, these prompts are tied to medical-imaging obligations: agents are reminded to preserve task-native data handling, avoid local tuning traps, inspect reference-backed directions before abandoning them, and maintain final-artifact evidence.

\section{Experiments}
\label{sec:experiments}
\subsection{Experimental Setup}

To evaluate \ourmethod{}'s ability to develop competitive medical imaging models, we use ReX-MLE~\citep{kenia2025rexmle}, a challenge benchmark with clear task definitions, standardized metrics, and competitive reference points. ReX-MLE contains 20 medical imaging tasks drawn from MICCAI-style and related evaluation settings. The suite covers segmentation, detection, classification, image-quality assessment, and enhancement across panoramic X-ray, CT, MRI, CTA, MRA, histopathology, ultrasound, and microscopy data. Each task provides public data, a task description, a scoring metric, and a required prediction or submission format, allowing us to assess whether an agent can build a valid model pipeline from raw files to final artifacts. Unless otherwise specified, the main experiments use the Codex backend with GPT-5.5 as the underlying coding agent. For each challenge, \ourmethod{} is given a 24-hour wall-clock budget and one NVIDIA RTX A6000 GPU with 48 GB of memory.

\subsection{Main Results and Comparative Analysis}

\begin{table}[!tbp]
\centering
\caption{Main ReX-MLE benchmark comparison across 20 medical imaging challenges. Challenge names and citations follow the ReX-MLE benchmark overview. All entries report accepted primary metric values in the original challenge units; Ours denotes \ourmethod{} with the Codex+GPT-5.5 backend, $\Delta$ reports the absolute improvement over the strongest listed agent baseline, and Human gives original-competition reference performance rather than a matched ReX-MLE split score.}
\label{tab:rexmle_tasks}
\small
\setlength{\tabcolsep}{4pt}
\renewcommand{\arraystretch}{1.05}
\resizebox{\linewidth}{!}{%
\begin{tabular}{llcccccc}
\toprule
\multicolumn{1}{c}{\multirow{2}{*}{\textbf{Challenge}}} & \multicolumn{1}{c}{\multirow{2}{*}{\textbf{Metric} ($\uparrow$)}} & \textbf{AIDE} & \textbf{ML-Master} & \textbf{R\&D-Agent} & \multirow{2}{*}{\textbf{Ours}} & \multirow{2}{*}{$\boldsymbol{\Delta}$} & \multirow{2}{*}{\textbf{Human}} \\
 & & {\scriptsize\citep{Jiang2025AIDEAE}} & {\scriptsize\citep{liu2025ml}} & {\scriptsize\citep{yang2025r}} & & & \\
\midrule
\multicolumn{8}{l}{\textbf{\textit{Segmentation Tasks}}} \\
ISLES'22~\citep{hernandez2022isles} & Dice & 0.04 & 0.00 & 0.02 & \textbf{0.71} & +0.67 & 0.79 \\
NeurIPS-CellSeg~\citep{neurips_cellseg} & F1 & 0.04 & 0.04 & 0.36 & \textbf{0.90} & +0.54 & 0.88 \\
PANTHER-T1~\citep{panther} & Dice & 0.33 & 0.13 & 0.16 & \textbf{0.42} & +0.09 & 0.73 \\
PANTHER-T2~\citep{panther} & Dice & 0.09 & 0.05 & 0.28 & \textbf{0.31} & +0.03 & 0.53 \\
PUMA-Track1-TissueSeg~\citep{puma} & Dice & \textcolor{red}{\textbf{FAIL}} & 0.00 & 0.00 & \textbf{0.56} & +0.56 & 0.78 \\
PUMA-Track2-TissueSeg~\citep{puma} & Dice & 0.00 & 0.00 & 0.00 & \textbf{0.56} & +0.56 & 0.78 \\
SEG.A~\citep{seg_a} & Dice & 0.02 & 0.02 & 0.00 & \textbf{0.91} & +0.89 & 0.92 \\
TopBrain-CTA-Seg~\citep{topcow} & Mean Dice & 0.03 & 0.26 & 0.08 & \textbf{0.59} & +0.33 & 0.79 \\
TopBrain-MRA-Seg~\citep{topcow} & Mean Dice & 0.01 & 0.26 & 0.50 & \textbf{0.64} & +0.14 & 0.81 \\
TopCoW-CTA-Seg~\citep{topcow} & Mean Dice & 0.09 & 0.25 & 0.49 & \textbf{0.63} & +0.14 & 0.87 \\
TopCoW-MRA-Seg~\citep{topcow} & Mean Dice & 0.11 & 0.48 & 0.73 & \textbf{0.76} & +0.03 & 0.88 \\
\midrule
\multicolumn{8}{l}{\textbf{\textit{Detection Tasks}}} \\
DENTEX~\citep{hamamci2023dentex} & AP & 0.09 & 0.08 & 0.09 & \textbf{0.49} & +0.40 & 0.40 \\
PUMA-Track1-NucleiDet~\citep{puma} & F1 & 0.02 & 0.08 & 0.06 & \textbf{0.54} & +0.46 & 0.66 \\
PUMA-Track2-NucleiDet~\citep{puma} & F1 & \textcolor{red}{\textbf{FAIL}} & 0.00 & 0.01 & \textbf{0.28} & +0.27 & 0.27 \\
TopCoW-CTA-Det~\citep{topcow} & IoU & 0.67 & 0.65 & 0.70 & \textbf{0.71} & +0.01 & 0.79 \\
TopCoW-MRA-Det~\citep{topcow} & IoU & 0.66 & 0.69 & 0.19 & \textbf{0.74} & +0.05 & 0.85 \\
\midrule
\multicolumn{8}{l}{\textbf{\textit{Classification Tasks}}} \\
TopCoW-CTA-Cls~\citep{topcow} & Accuracy & 0.33 & 0.10 & 0.28 & \textbf{0.33} & +0.00 & 0.73 \\
TopCoW-MRA-Cls~\citep{topcow} & Accuracy & 0.33 & 0.33 & 0.09 & \textbf{0.46} & +0.13 & 0.89 \\
\midrule
\multicolumn{8}{l}{\textbf{\textit{Image Quality \& Enhancement Tasks}}} \\
LDCT-IQA~\citep{ldct_iqa} & Score & 2.62 & 2.50 & 2.66 & \textbf{2.74} & +0.08 & 2.74 \\
USenhance~\citep{usenhance} & LNCC & 0.11 & 0.13 & \textcolor{red}{\textbf{FAIL}} & \textbf{0.19} & +0.06 & 0.91 \\
\bottomrule
\end{tabular}}
\end{table}

Table~\ref{tab:rexmle_tasks} gives the main comparison across the full 20-task ReX-MLE suite. We compare \ourmethod{} with the three autonomous MLE systems reported in ReX-MLE: AIDE~\citep{Jiang2025AIDEAE}, ML-Master~\citep{liu2025ml}, and R\&D-Agent~\citep{yang2025r}. To keep the comparison focused on system design rather than metric reinterpretation, the baseline columns use the accepted benchmark scores reported by ReX-MLE under its original challenge contracts, while \ourmethod{} is evaluated with the same accepted-artifact criterion. The comparison also uses GPT-family backends on both sides: the ReX-MLE baselines use GPT as their primary backend, and our main \ourmethod{} run uses Codex with GPT-5.5. Thus, each entry reflects an end-to-end result, not a local validation number that might be invalidated by prediction format, metric direction, split discipline, or final submission packaging. The Human column is included only as a reference scale. These values are the original competition reference scores obtained on the true challenge test sets, which are not publicly released. ReX-MLE therefore constructs its evaluation by re-splitting the originally public training data; agents in this setting train with less data than teams had in the original competitions. Human-normalized comparisons should consequently be read as approximate performance references rather than strictly matched leaderboard comparisons.

\begin{figure}[!tbp]
\noindent\makebox[\linewidth][l]{\includegraphics[width=0.96\linewidth]{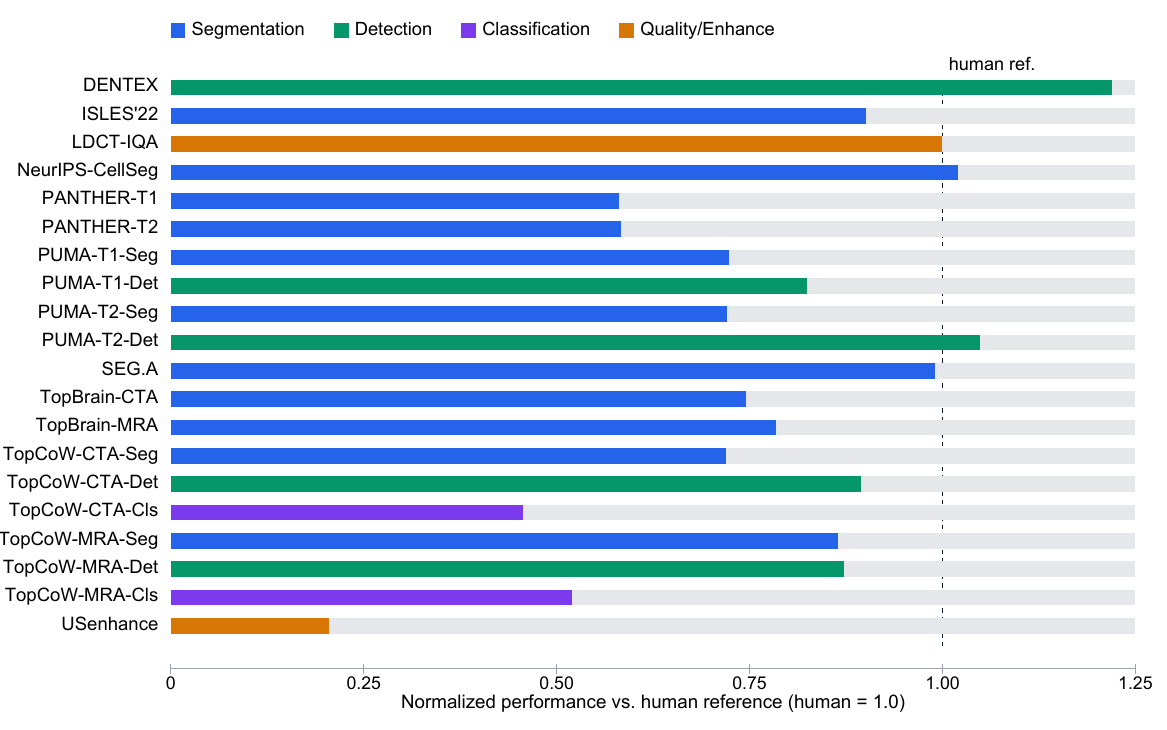}}
\caption{Challenge-wise comparison with human reference performance on ReX-MLE. Bars show accepted \ourmethod{} performance normalized by each challenge's original-competition human reference; the dashed line marks reference parity and is used as an approximate scale rather than as a strictly matched leaderboard target.}
\label{fig:codex_sheet3_results}
\end{figure}

Across the 20 challenges, \ourmethod{} produces valid accepted results for every task and improves over the strongest listed baseline on 19 tasks, with the remaining task tied on TopCoW-CTA-Cls. The gains are most pronounced on segmentation and detection tasks, where completing the medical-imaging pipeline is often the main bottleneck for general-purpose agents. In segmentation, \ourmethod{} raises SEG.A Dice by +0.89 and ISLES'22 Dice by +0.67 over the best baseline, while also converting previously failed or near-zero PUMA tissue-segmentation runs into valid Dice scores above 0.5. In detection, \ourmethod{} improves DENTEX AP by +0.40 and the PUMA-T1 nuclei detection F1 score by +0.46. Relative to the human reference scale in Figure~\ref{fig:codex_sheet3_results}, the strongest \ourmethod{} results reach or exceed the reference on DENTEX, NeurIPS-CellSeg, PUMA-T2-Det, and LDCT-IQA, and come close on SEG.A. Performance remains weaker on graph classification, vessel-analysis tasks that depend on small anatomical structures, and ultrasound enhancement. These gaps show that \ourmethod{} is broadly stronger than existing autonomous MLE baselines, but expert-level performance remains task-dependent rather than automatic.

\begin{table}[!tbp]
\centering
\caption{Focused comparison on three representative ReX-MLE challenges. All metrics are higher-is-better. The first three rows report agent baselines under a unified GPT-5.5 model backend; the final two rows report \ourmethod{} under Codex+GPT-5.5 and Claude Code runtime settings for the same challenge definitions.}
\label{tab:backend_comparison}
\small
\setlength{\tabcolsep}{2.5pt}
\renewcommand{\arraystretch}{1.05}
\begin{tabular}{@{}lcccc@{}}
\toprule
\multicolumn{1}{c}{\multirow{2}{*}{\textbf{Method}}} & \multirow{2}{*}{\textbf{Model backend}} & \textbf{DENTEX} & \textbf{PUMA-T1-Seg} & \textbf{TopCoW-MRA-Cls} \\
 & & \textbf{AP} ($\uparrow$) & \textbf{Dice} ($\uparrow$) & \textbf{Acc.} ($\uparrow$) \\
\midrule
AIDE~\citep{Jiang2025AIDEAE} & GPT-5.5 & 0.08 & 0.39 & \textcolor{red}{\textbf{FAIL}} \\
ML-Master~\citep{liu2025ml} & GPT-5.5 & 0.06 & 0.50 & 0.39 \\
R\&D-Agent~\citep{yang2025r} & GPT-5.5 & 0.08 & 0.52 & 0.09 \\
\midrule
Ours (Codex) & GPT-5.5 & \textbf{0.49} & 0.56 & 0.46 \\
Ours (Claude Code) & Opus-4.8 & 0.25 & \textbf{0.64} & \textbf{0.50} \\
\bottomrule
\end{tabular}
\end{table}

\paragraph{Unified model-backend comparison.}
Table~\ref{tab:backend_comparison} complements the full sweep with a controlled view on three representative challenges. The first three rows hold the underlying model backend fixed at GPT-5.5 for AIDE, ML-Master, and R\&D-Agent. Even under this stronger and unified model setting, the best baseline scores remain below the best \ourmethod{} row on each representative task: 0.08 AP versus 0.49 AP on DENTEX, 0.52 Dice versus 0.64 Dice on PUMA-T1-Seg, and 0.39 accuracy versus 0.50 accuracy on TopCoW-MRA-Cls. This comparison rules out a weak-baseline explanation for the focused gaps and suggests that the higher-level search organization, evidence tracking, and final-artifact verification are important beyond the model backend alone.

\paragraph{Runtime-backend behavior within \ourmethod{}.}
The final two rows of Table~\ref{tab:backend_comparison} show \ourmethod{} under different model and runtime configurations. The Codex+GPT-5.5 row uses the same model backend and runtime family as the main full-suite run, while the Claude Code row probes the same \ourmethod{} search layer under a different runtime and model backend. Codex+GPT-5.5 is strongest on DENTEX, reaching 0.49 AP, while Claude Code obtains the best focused scores on PUMA-T1-Seg and TopCoW-MRA-Cls, with 0.64 Dice and 0.50 accuracy. This mixed pattern indicates that runtime backends still affect individual implementation trajectories, debugging behavior, and task-specific method choices. However, the consistent advantage of the \ourmethod{} rows over the corresponding baseline rows under the same challenge definitions supports the main claim: \ourmethod{}'s gains come from organizing and verifying the model-building process, not from a single favorable runtime alone.

\FloatBarrier
\subsection{Case Studies}

\paragraph{Strong cases: search and task-specific adaptation.}
In this section, we analyze why \ourmethod{} performs well in representative strong cases: broad search identifies task-matched methods, task-specific adaptation converts them into valid challenge solutions, and artifact verification keeps the final outputs aligned with the submission contract. DENTEX is a representative example. The final submission does not rely on a detector alone: it starts from a compact YOLOv8~\citep{yolov8} detector adapted to panoramic dental radiographs, then adds an anatomical calibration layer that maps unordered boxes to FDI quadrants and tooth numbers using dental-arch templates, conservative diagnosis fallback statistics, and learned alternate tooth-position rules. This hybrid design reaches 0.48825 mean AP, 0.64702 AP50, and 0.68248 AR, outperforming the autonomous-agent baselines in Table~\ref{tab:rexmle_tasks}. The gain comes from matching the task structure: tooth enumeration is partly visual detection and partly anatomical ordering, so the selected solution combines a learned detector with structured domain priors instead of treating all labels as independent detector classes.

The PUMA pathology tasks show a second form of the same behavior. For tissue segmentation, \ourmethod{} discovers pathology foundation encoders and uses them as the stable representation layer: PUMA Track~1 uses a locked five-fold ensemble over frozen UNI~\citep{uni} patch-token features, where lightweight MLP heads combine local tissue tokens with RGB and coordinate cues; PUMA Track~2 fine-tunes the final Phikon-v2~\citep{filiot2024phikon} block and blends full-image, fixed-crop, and overlapping-crop evidence. These choices convert tasks where prior autonomous agents failed or scored zero into accepted Dice scores of 0.56395 and 0.56214. For nuclei detection, the system chooses a different decomposition. PUMA Track~1 first uses CellViT-256~\citep{horst2024cellvit} to propose nuclei and token embeddings, then trains a deterministic gradient-boosting classifier with local and neighborhood context; PUMA Track~2 uses PathoSAM~\citep{pathosam} proposals and Midnight pathology embeddings with a small classical classifier ensemble. The resulting macro-F1 scores, 0.54359 for the three-class task and 0.28422 for the ten-class task, are especially informative because the Track~2 result ranks first in the official track despite a low absolute macro-F1 caused by rare cell classes. In both cases, broad method search matters: the selected pipelines are built around domain foundation models and task-specific post-processing, not around generic CNN training from scratch.

The broader sweep supports the same interpretation. On NeurIPS-CellSeg, \ourmethod{} fine-tunes Cellpose-SAM~\citep{cellposesam} and reaches a 0.893655 internal accepted score with official F1 values near 0.90 at IoU 0.5. On SEG.A and ISLES'22, it falls back to strong task-appropriate medical segmentation baselines, using nnU-Net-style~\citep{isensee2021nnu} planning, late-checkpoint selection, and conservative anatomical cleanup or recall-oriented mask unions. On TopCoW detection, it uses geometry priors, heatmap center prediction, segmentation-derived boxes, and lightweight direct detectors rather than forcing one generic detector across CTA and MRA. These traces illustrate why the two-stage optimization is useful: exploration supplies several plausible method lanes, while exploitation and artifact verification select the lane that both scores well and satisfies the exact submission contract.

\paragraph{Failure study.}
USenhance highlights a limitation of long-horizon image translation rather than a simple implementation error. The submitted five-fold pix2pix-style~\citep{pix2pix} U-Net ensemble is complete and produces valid outputs, but its official result remains low, with LNCC 0.18658, SSIM 0.38495, and PSNR 17.12416. Stronger CycleGAN-style~\citep{cyclegan} adversarial candidates were difficult to promote because generator-discriminator losses were unstable, and early loss curves converged slowly. Under a fixed budget, these candidates therefore received weak short-horizon evidence and were rejected in favor of the more stable paired U-Net recipe, even though the challenge likely needs longer, carefully monitored training to balance anatomical fidelity, speckle realism, and local-normalized-correlation objectives.

The two TopCoW classification tasks expose a coupled segmentation-classification bottleneck. Although the official target is graph-edge classification, useful labels depend on an intermediate vessel segmentation and topology extraction for small structures such as PCom, ACom, and a third A2 segment. The CTA run collapses to a fixed topology prior, while the MRA run uses SegResNet vessel masks and contact rules but still reaches only 0.46296 anterior accuracy and 0.16667 posterior accuracy. This contrasts with the stronger pure TopCoW segmentation runs, suggesting that voxel-level vessel masks are not enough when the downstream objective depends on small branch contacts and graph labels. These tasks require segmentation, topology post-processing, and classification-oriented selection to improve together, making them harder than isolated segmentation or detection under the same budget. 

We make the solutions obtained for all challenges publicly available on the project page.

\section{Discussion}
\label{sec:discussion}
\paragraph{Medical model development as a search problem.}
\ourmethod{} treats autonomous medical-imaging model development as a search problem over executable and verifiable solution candidates rather than as free-form idea generation or a single prompt-to-code task. This distinction is important because a useful candidate in medical imaging is more than a plausible architecture; it must match the modality, anatomy, annotation semantics, patient or case organization, metric direction, data scale, compute budget, and final artifact contract. Data-Conditioned Method Planning therefore turns the task and data profile into a portfolio of method lanes, where each lane carries a modeling hypothesis, implementation resources, feasibility assumptions, cost expectations, validation obligations, and likely failure modes. In this sense, \ourmethod{}'s ``idea generation'' step is deliberately closer to a domain-conditioned search system: it expands multiple medically plausible routes, downranks paper-only or incompatible resources, and makes the candidate space inspectable before expensive experimentation.

\paragraph{Verification as the control layer.}
The results also suggest that autonomous medical MLE should be evaluated based on verified evidence rather than on whether generated code happens to run. A runnable script can still use the wrong split, optimize the wrong metric, leak patient-level information, write malformed prediction files, duplicate a previous payload, or report a score that cannot be traced to a final submission. \ourmethod{} addresses this failure mode by attaching acceptance criteria to each attempt: validation scores must be linked to committed code, logs, fold or protocol records, metric computation, prediction artifacts, and lane attribution. In both stages, reviewer checks act as an internal validity layer, checking metric direction, legal data access, fold provenance, complete outputs, prediction schema, and traceability from accepted attempts to final packages. This verification layer is not only a final safety check after optimization. It changes the optimization itself, because candidates that cannot satisfy the protocol are used for diagnosis or repair but are not allowed to drive promotion or final selection.

\paragraph{From fine-grained prompting to agent-loop orchestration.}
\ourmethod{} also reflects a shift in how long-horizon agents can be engineered. Earlier agent systems often relied on finely designed prompts for individual steps such as planning, coding, debugging, and summarization. Modern coding-agent runtimes already provide much of this execution harness: persistent workspaces, terminals, version control, tool use, and iterative repair. The remaining design problem is therefore less about hand-writing every low-level instruction and more about building the higher-level substrate around the harness. In \ourmethod{}, this substrate consists of the task contract, shared file-system memory, method-lane records, attempt ledgers, heartbeat interventions, reusable skills, lifecycle scheduling, and final-artifact gates. These components let heterogeneous coding agents operate as workers inside the same auditable search process, while the system-level loop decides which lanes need coverage, which candidates merit exploitation, and which claims are supported by preserved evidence.

\paragraph{Scope beyond challenge submissions.}
The challenge benchmark setting is useful because it provides clear metrics, standardized artifacts, and competitive reference points, but the same design is not tied to challenge participation. The target deliverable can be a submission file, a trained model package, a Dockerized inference pipeline, a prediction service, or an evidence bundle for human review. Fixed-time-budget evaluation is one reproducible deployment mode; open-ended use can instead allow users to inspect evidence, stop a run, request additional exploration, or select among credible candidates. At the same time, challenge performance should be interpreted as system-level evidence rather than as clinical validation. It measures task-specific predictive performance under a defined evaluation protocol, but it does not establish clinical safety, prospective generalization, calibration under deployment shift, or suitability for patient care. \ourmethod{}'s artifact-first design is meant to make such later validation easier by preserving how a model was built, which resources it used, and where remaining risks may enter.

\paragraph{Limitations and next steps.}
The main limitation is cost. Long-horizon medical MLE consumes both GPU time and large numbers of coding-agent tokens, especially when the system explores multiple method lanes, runs stage-level reviewer checks, and preserves evidence across workers. For this reason, our current experiments do not exhaustively evaluate every challenge under all of the latest model-runtime combinations; the full-suite results use the primary backend, while newer or alternative backends are examined on focused representative tasks. Broader sweeps across model backends, compute budgets, method-lane ablations, verification ablations, and reviewer-style validity analyses remain necessary to separate the contribution of the search design from the strength of the underlying agent runtime. \ourmethod{} also depends on the quality of the resource registry, the accuracy of data profiling, and the backend agent's ability to implement strong methods under budget. A key research direction is therefore cost-aware long-horizon autonomy: using cheaper early filters, stronger stopping rules, reusable skill distillation, cached resource audits, surrogate validation, and selective escalation to expensive models only when a candidate is worth deeper verification.

\section{Related Work}
\label{sec:related}
\paragraph{Autonomous empirical software and MLE agents.}
Autonomous empirical software agents define progress through executable artifacts rather than through static text generation. Expert-level empirical software writing~\citep{aygun2026ai}, code-space exploration~\citep{Jiang2025AIDEAE}, and AI-driven systems research~\citep{cheng2025barbarians} all rely on an execution-feedback loop in which code, logs, tests, and validation scores become the evidence for subsequent decisions. SWE-bench~\citep{jimenez2024swebench} formalizes execution-feedback evaluation for repository-level software repair, whereas MLE-bench~\citep{chan2024mle} and MLAgentBench~\citep{huang2024mlagentbench} formalize autonomous model development as the production of training code, validation evidence, debugging traces, and final submissions. Recent benchmarks extend executable evaluation from general model building to specialized long-horizon AI engineering settings. PostTrainBench~\citep{rank2026posttrainbench} focuses on autonomous LLM post-training under bounded compute, while MLS-Bench~\citep{lyu2026mlsbench} evaluates whether agents can propose, test, and validate ML improvements that generalize beyond a single tuned pipeline. Within benchmarked MLE settings, recent agent frameworks increasingly represent model development as structured search over runnable pipelines. ML-Master~\citep{liu2025ml} uses MCTS-style search to maintain competing solution branches, while AutoMLGen~\citep{du2025automlgen} extends branch-level search with graph structure, intra-branch evolution, cross-branch reference, and multi-branch aggregation so that partial findings can propagate across the search graph. MLEvolve~\citep{du2026mlevolve} translates graph/tree search into specialized draft, improvement, debugging, evolution, fusion, result-parsing, and code-review agents, while Arbor~\citep{jin2026toward} stores hypotheses, evidence, branches, and distilled insights in a persistent hypothesis tree. Recent studies~\citep{zhang2026aibuildai,qu2026coral,gao2026autoscientists} further treat general coding-agent runtimes, such as Claude Code or Codex, as substrates for higher-level autonomous systems, using runtime-backed repositories, shared persistent memory, heartbeat interventions, and self-organizing teams to support long-running experimentation.

\paragraph{Autonomous research systems.}
Autonomous research systems generalize empirical software loops into end-to-end scientific workflows encompassing hypothesis generation, literature grounding, experiment design and execution, result interpretation, manuscript drafting, and review. A growing body of research-agent benchmarks~\citep{starace2025paperbench,kim2025autoexperiment,siegel2024corebench,hu2025reprobench,ye2025replicationbench,huang2026automat,faroughy2026colliderbench,wang2026firebench} has operationalized this agenda through paper-grounded reproduction, replication, workflow reconstruction, and insight rediscovery tasks, revealing persistent limitations in faithful execution, long-horizon experiment management, and evidence-grounded reasoning that motivate more structured autonomous research systems. In parallel, early studies of autonomous scientific research and tool-augmented scientific reasoning~\citep{boiko2023emergent,wang2024sciagent} showed that language-model agents can coordinate scientific actions beyond static question answering. Building on this premise, The AI Scientist~\citep{lu2024aiscientist} provided an early end-to-end formulation of idea-to-paper automation by mapping research ideas into experiments and manuscripts within constrained templates, while AI Scientist-v2~\citep{yamada2025aiscientistv2} strengthened this formulation through agentic tree search and review-aware iteration. This line of work has since expanded the design space along several complementary dimensions: AI co-scientist~\citep{gottweis2025coscientist} emphasizes multi-agent hypothesis generation, whereas CodeScientist~\citep{jansen2025codescientist} grounds scientific discovery jointly in literature and executable code; related systems~\citep{wu2025piflow,ifargan2024data2paper,schmidgall2025agentlab,tang2025airesearcher,huang2025deepscientist} explore principle-guided exploration, evidence-aware manuscript generation, human-in-the-loop assistance, and autonomous research ideation. Another strand addresses the operational bottlenecks of idea-to-paper pipelines by equipping agents with more durable execution substrates: MARS~\citep{chen2026mars} and AutoSOTA~\citep{li2026autosota} frame empirical AI research as repository-level optimization over papers, code, dependencies, compute budgets, reusable skills, and reproducible performance gains, while ScientistOne~\citep{meng2026scientistone} elevates claim-level provenance to a first-class design constraint by tracking chains of evidence across literature, experiments, code, and writing. Complementing these efforts, AutoResearchClaw~\citep{li2024autoresearchclaw} and ARIS~\citep{yang2026aris} incorporate reusable skills directly into the research state, enabling agents to retrieve, adapt, and accumulate procedures for experiment repair, evidence checking, and manuscript construction rather than relying on one-off prompting or fixed pipelines. Taken together, these developments mark a shift from isolated demonstrations of scientific agency toward infrastructure-aware, provenance-grounded, and skill-accumulating autonomous research systems capable of sustaining iterative scientific workflows.

\paragraph{Medical autoresearch systems.}
Medical autoresearch systems can lower the barrier for clinicians to translate clinical or imaging questions into usable models, analyses, and manuscript-ready evidence, provided that they produce runnable development workflows and auditable artifacts rather than clinical answers alone. ReX-MLE~\citep{kenia2025rexmle} makes this requirement explicit by evaluating agents on medical-imaging challenge tasks that require preprocessing, model development, metric handling, and valid submissions. AutoMedBench~\citep{liu2026automedbench} and HealthAgentBench~\citep{liu2026healthagentbench} further demonstrate that current agents still struggle with end-to-end clinical workflows that combine large search spaces with compositional reasoning. To solve these problems, medical AI scientist systems~\citep{wu2026towards} and Camyla~\citep{camyla2026} extend autonomous research into biomedical discovery and medical image segmentation, while related neuroimaging and pathology systems~\citep{wang2026neuroclaw,han2026nexus,trost2026agentic} emphasize domain-specific scientific assistance, analysis, or discovery workflows rather than general medical-imaging model construction. A recent study~\citep{zhao2026clinical} uses coding agents to translate natural-language clinical requests into runnable development loops and refine models under clinician-specified priorities. However, its demonstrations remain limited to a small set of constrained tasks and leave open how an agent should optimize across competing modeling strategies and how the selected result should be verified before delivery. \ourmethod{} focuses on this optimization-and-verification problem through Verification-Guided Two-Stage Optimization and artifact auditing, so model packages are promoted, rejected, or accepted based on executable evidence rather than clinician steering alone.

\section{Conclusion}
\label{sec:conclusion}
In this technical report, we presented \ourmethod{}, an autonomous multi-agent framework for verifiable medical imaging AI development. \ourmethod{} formulates model building as a data-conditioned and verification-controlled search process: it establishes the task contract, profiles the medical-imaging data, constructs executable method lanes, coordinates broad exploration and selective exploitation, and accepts final outputs only when reviewer-checked evidence and artifacts satisfy the required protocol. Across the ReX-MLE medical-imaging challenge suite, this design improves over existing autonomous MLE baselines on most evaluated tasks and reaches expert-level performance on selected challenges, suggesting that domain-conditioned planning and artifact-level verification are important ingredients for reliable autonomous medical MLE.

This manuscript should be read as an initial technical report rather than a definitive claim about a finished system. The current version establishes the framework, reports the first benchmark evidence, and exposes the design choices needed to make agentic medical model development auditable. We will continue to improve \ourmethod{} by expanding controlled backend comparisons, adding ablations for data profiling, method-lane planning, two-stage optimization, and verification gates, strengthening reviewer-style audit analyses, and evaluating cost-aware strategies for long-horizon runs. Future versions will also broaden validation across additional tasks, model-runtime combinations, resource settings, and deployment-oriented artifact checks, so that the system's performance gains, failure modes, and practical boundaries can be characterized more completely.

\bibliographystyle{plainnat}
\bibliography{references}

\end{document}